\documentclass[acmtog, nonacm, timestamp]{acmart}
\authorsaddresses{}
\renewcommand\footnotetextcopyrightpermission[1]{}

\usepackage{booktabs}  
\usepackage{multirow}  
\usepackage{graphicx}  
\usepackage[table]{xcolor} 
\usepackage{xcolor}

\usepackage{enumitem}
\setlist[itemize]{leftmargin=*}

\AtBeginDocument{%
  }

\acmDOI{XX.XX}

\begin{document}

\title{Flatten The Complex: Joint B-Rep Generation via Compositional $k$-Cell Particles}

\author{Junran Lu}
\email{junranlu@smail.nju.edu.cn}
% \orcid{0009-0007-7595-8984}
\affiliation{%
  \institution{Nanjing University}
  % \city{Nanjing}
  % \state{Jiangsu}
  \country{China}
}

\author{Yuanqi Li}
\email{yuanqili@nju.edu.cn}
% \orcid{0000-0003-4100-7471}
\affiliation{%
  \institution{Nanjing University}
  % \city{Nanjing}
  % \state{Jiangsu}
  \country{China}
}

\author{Hengji Li}
\email{221240073@smail.nju.edu.cn}
% \orcid{0009-0006-5368-687X}
\affiliation{%
  \institution{Nanjing University}
  % \city{Nanjing}
  % \state{Jiangsu}
  \country{China}
}

\author{Jie Guo}
\email{guojie@nju.edu.cn}
% \orcid{0000-0002-4176-7617}
\affiliation{%
  \institution{Nanjing University}
  % \city{Nanjing}
  % \state{Jiangsu}
  \country{China}
}

\author{Yanwen Guo}
\email{ ywguo@nju.edu.cn}
% \orcid{0000-0002-7605-5206}
\affiliation{%
  \institution{Nanjing University}
  % \city{Nanjing}
  % \state{Jiangsu}
  \country{China}
}

\begin{abstract}
  Boundary Representation (B-Rep) is the widely adopted standard
  in Computer-Aided Design (CAD) and manufacturing. However, generative modeling of B-Reps remains a formidable challenge due to their inherent heterogeneity as geometric cell complexes, which entangles topology with geometry across cells of varying orders (\textit{i.e.}, $k$-cells such as vertices, edges, faces). Previous methods typically rely on cascaded sequences to handle this hierarchy, which fails to fully exploit the geometric relationships between cells, such as adjacency and sharing, limiting context awareness and error recovery. To fill this gap, we introduce a novel paradigm that reformulates B-Reps into sets of compositional k-cell particles. Our approach encodes each topological entity as a composition of particles, where adjacent cells share identical latents at their interfaces, thereby promoting geometric coupling along shared boundaries. By decoupling the rigid hierarchy, our representation unifies vertices, edges, and faces, enabling the joint generation of topology and geometry with global context awareness.
  We synthesize these particle sets using a multi-modal flow matching framework to handle unconditional generation as well as precise conditional tasks, such as 3D reconstruction from single-view or point cloud. Furthermore, the explicit and localized nature of our representation naturally extends to downstream tasks like local in-painting and enables the direct synthesis of non-manifold structures (\textit{e.g.}, wireframes). Extensive experiments demonstrate that our method produces high-fidelity CAD models with superior validity and editability compared to state-of-the-art methods.

\end{abstract}




\begin{teaserfigure}
\vspace{-3mm}
  \includegraphics[width=\textwidth]{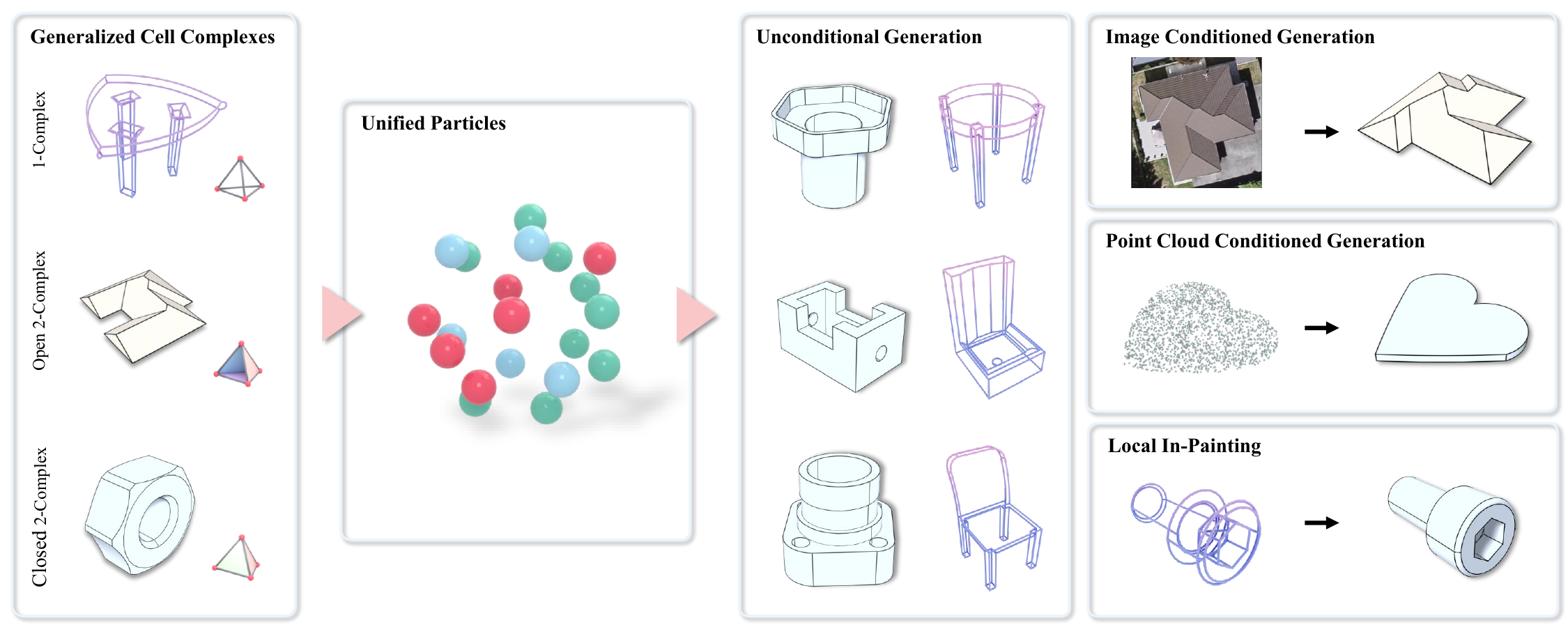}
  \vspace{-6mm}
  \Description{todo}
  \caption{Our method introduces a particle-based representation that unifies topological entities. This enables the joint generation of geometry and topology with global awareness, supporting tasks from unconditional shape synthesis to conditional reconstruction and local editing.}
  \label{fig:teaser}
  %\vspace{-1mm}
\end{teaserfigure}

\maketitle

% Include sections
\section{Introduction}
\label{sec:intro}

Boundary Representation (B-Rep) models constitute the foundational language of modern engineering. Unlike meshes or implicit fields, B-Reps rigorously encode 3D shapes as geometric cell complexes, inextricably intertwining continuous parametric surfaces with discrete topological relations. While this structure enables the infinite precision required for CAD/CAM, it presents a formidable challenge for generative modeling. Modern generative models excel at homogeneous, grid-like, or set-like data, such as meshes~\cite{PolyDiff, PolyGen}, implicit fields~\cite{vectorset, LanHYZMDPL24}, point cloud~\cite{LION}, voxels~\cite{trellis}. In contrast, the heterogeneous and strictly constrained nature of B-Reps creates a structural mismatch, rendering them resistant to direct synthesis via these prevailing 3D generative paradigms.

Driven by this complexity, prior works primarily rely on progressive synthesis. One prevailing paradigm, exemplified by SolidGen~\cite{SolidGen} and BrepGen~\cite{BrepGen}, formulates generation as a sequential or cascaded process. Specifically, SolidGen linearizes the B-Rep for autoregressive modeling, while BrepGen employs a hierarchical cascade to synthesize entities in a coarse-to-fine manner. However, such step-wise dependency hinders global context reasoning and inevitably leads to error accumulation, where early geometric inaccuracies cascade into downstream topological failures. Conversely, holistic approaches attempt to generate entities simultaneously to capture global coherence.
Notably, ComplexGen~\cite{ComplexGen} advances this direction by jointly reconstructing the full complex; however, it treats topological entities primarily as abstract graph nodes with attached latent attributes. 
Crucially, this abstraction neglects the spatial embedding nature of B-Reps: without explicit spatial grounding, such methods rely on soft attention mechanisms to model boundary constraints, lacking the structural guarantees necessary for precise geometric alignment.

In this work, we propose to transcend these limitations by fundamentally reformulating B-Reps. We observe that B-Reps are not merely abstract topological graphs, but are strictly spatially embedded. Topological connectivity is strongly correlated with Euclidean proximity, suggesting that discrete relations can be effectively compressed into continuous, spatially localized attributes.
Based on this insight, we introduce Compositional $k$-Cell Particles (KCPs), a novel representation that flattens the hierarchical cell complex into a unified, unordered set of spatially anchored feature vectors.
To resolve the detached geometry problem, we introduce a compositional sharing mechanism: high-order entities do not define their boundaries independently but structurally reuse the particle features of their lower-order neighbors. By structurally sharing features at spatially anchored interfaces, our method introduces a strong inductive bias for geometric continuity. This design treats B-Reps not as abstract graphs, but as coherent particle fields, naturally promoting seamless alignment by exploiting spatial locality.

By shifting the inherent complexity of B-Reps from the architecture into this compositional representation, we establish a streamlined, end-to-end generative pipeline. This approach transforms the synthesis of heterogeneous graphs into a scalable set generation task, leading to the following contributions:

\begin{itemize}[nosep]
\item \textbf{Compositional $k$-Cell Particles (KCPs).} We introduce a novel B-Rep representation that unifies heterogeneous $k$-cells into a set of spatially anchored particles. By imposing a compositional decoding mechanism where boundary features are shared, we inject a strong inductive bias for geometric continuity.
\item \textbf{Holistic Parallel Generation via Flow Matching.} We propose a scalable set-based generative framework that treats B-Rep synthesis as a set prediction problem, enabling holistic global reasoning. Our method supports multi-modal conditional generation using image or point cloud inputs and outperforms state-of-the-art methods in generating valid, complex B-Reps.
\item \textbf{Spatial Editability \& Generalization.} The spatially anchored nature of our representation facilitates seamless adaptation to versatile downstream tasks, such as partial B-Rep completion and non-manifold synthesis, without architectural modifications.
\end{itemize}

\section{Related Work}
\label{sec:related_work}

\subsection{Representation Learning on B-Rep}
Modeling B-Reps as heterogeneous graphs is a theoretically grounded approach that combines discrete topology with continuous geometry~\cite{AnsaldiFF85}.

Pioneering works in B-Rep analysis have validated this direction. UVNet \cite{UVNet} extracts features from B-Rep face adjacency graphs for classification. BRepNet \cite{BrepNet} and subsequent works like AutoMate~\cite{AutoMate} define convolution operations explicitly over the topological hierarchy, treating the assembly of entities as a rigorous message-passing process.

However, these architectures act primarily as an analytic method for topology; adapting them for generative tasks is non-trivial, due to their complex hierarchical network design.

\subsection{Indirect B-Rep Generative Modeling}
To circumvent the complexity of directly synthesizing heterogeneous graphs, many approaches resort to indirect representations.

Sequence-based methods treat shape creation as a linear stream of commands \cite{DeepCAD, CADCrafter} or executable code generated by LLMs \cite{GaninBLKS21, CADRecode, AlrashedyTZLXG25, WuKKJPWL24}. Similarly, some works focus on inferring construction histories \cite{Xu2021InferringCM, LiPBM22} or Constructive Solid Geometry (CSG) trees \cite{Sharma2017CSGNetNS, kania2020ucsg, Ren2021CSGStumpAL, Yu2023D2CSGUL, RitchieGJMSWW23} to capture design intent.

Alternatively, proxy-based methods generate intermediate representations like voxel grids \cite{ReconstructingEP}, Gaussian Splats \cite{CADDreamer}, or 2D drawings \cite{Text2CAD} for subsequent reconstruction.
Nevertheless, these methods are constrained: sequence and code-based models lack topological flexibility, while proxy-based methods often suffer from conversion artifacts.

\subsection{Direct B-Rep Generative Modeling}
Recent advances have shifted towards directly synthesizing B-Rep data structures, which can be broadly categorized into hierarchical and holistic frameworks.

\paragraph{Hierarchical Frameworks.}
To ensure structural validity, these methods decompose B-Reps into a strict hierarchy or sequence.
BrepGen \cite{BrepGen} and DTGBrepGen \cite{DTGBrepGen} predict topology and geometry via a structured top-down cascade (e.g., binary trees), while SolidGen \cite{SolidGen}, AutoBrep \cite{AutoBrep}, and Stitch-A-Shape \cite{StitchAShape} utilize pointer networks to generate entities element-by-element.

However, a critical limitation of these approaches is \textit{error accumulation}: since generation is sequential, early geometric deviations propagate downstream, often resulting in broken loops or invalid faces. Moreover, the lack of a holistic view restricts global context awareness, making local editing tasks non-trivial without re-generating the entire sequence.

\paragraph{Holistic Approaches.}
To mitigate sequential errors, recent research explores parallel or single-stage frameworks.
Strategies in this category vary in scope.

Some methods simplify the challenge by prioritizing a single $k$-cell type: Like face-centric \cite{HoLa, BrepDiff, BrepGiff}, edge-centric~\cite{CLRWire} or \mbox{point-centric~\cite{StitchAShape}}. Such approaches typically depend on ad-hoc heuristics or post-inference networks to complete the shape, lacking a unified formulation.

Conversely, ComplexGen \cite{ComplexGen} advances the field by jointly reconstructing the full complex. 
While it captures the complete graph structure, it represents topological entities as abstract latent vectors lacking explicit spatial structure. It neglects the inherent spatial embedding of B-Reps, thereby treating geometry as isolated attributes. In contrast, our approach exploits spatial locality: by ensuring topological neighbors share proximate features, we inherently foster geometric alignment without relying on fragile soft attention.

\begin{figure*}[t]
    \centering
    \includegraphics[width=\linewidth]{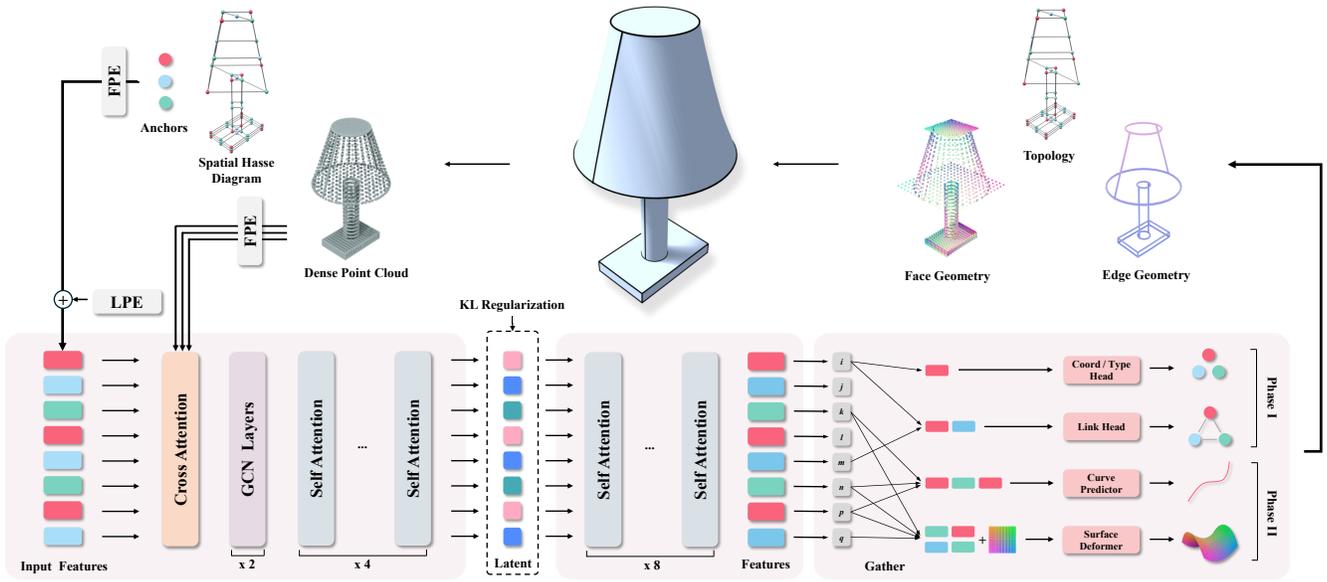}
    \Description{Detailed architecture of the CC-VAE.}
    \caption{Architecture of the CC-VAE. The framework encodes a B-Rep into a set of latent particles and reconstructs it via a compositional decoder. 
\textbf{Left:} The encoder fuses local surface details extracted from a dense point cloud with topological context propagated along the input Spatial Hasse Diagram via GCN layers. A Transformer encoder aggregates these features into the latent particle set subject to KL regularization.
\textbf{Right:} The reconstruction proceeds in two stages. \textit{Phase I} recovers the Spatial Hasse Diagram by predicting particle attributes and incidence links. \textit{Phase II} performs compositional geometry synthesis, where specific geometry heads generate detailed geometry conditioned on features gathered via the connectivity recovered in Phase I.}
    \label{fig:VAE-architecture}
\end{figure*}

\begin{figure}[H]
    \centering
    \includegraphics[width=\linewidth]{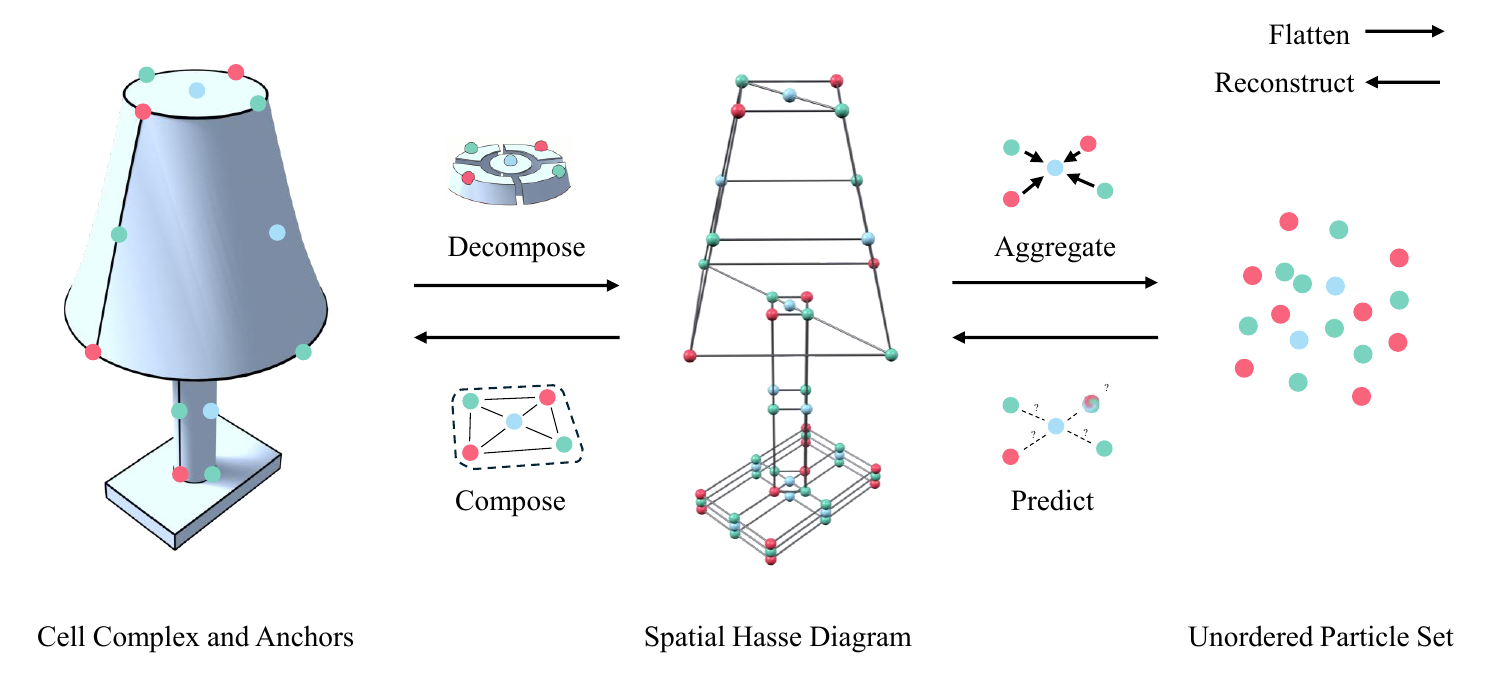}
    \Description{A diagram illustrating the transformation of a B-Rep into a set of particles and back. Left: Hierarchical B-Rep. Right: Unordered set of k-cell particles. The Flatten process maps cells to particles, aggregating geometry and topology. The Restore process uses compositional prediction to reconstruct watertight geometry.}
    \caption{Illustration of the reformulation of the hierarchical B-Rep model into an unordered set of compositional $k$-cell particles, and the subsequent reconstruction of the original hierarchical structure..}
    \vspace{-0.5em}
    \label{fig:particle_system}
\end{figure}

\section{Methodology}

We aim to generate B-Rep models that are topologically valid and geometrically precise without adhering to strict manifold assumptions. We reformulate B-Rep synthesis as a set generation problem by mapping the hierarchical cell complex into a Compositional $k$-Cell Particle (KCP) representation.

Our framework consists of two stages: (1) a \textbf{Compositional Cell VAE (CC-VAE)} (Sec.~\ref{sec:vae}) that learns a continuous latent space over the Spatial Hasse Diagram of the complex; and (2) a \textbf{Rectified Flow Transformer} (Sec.~\ref{sec:flow}) that models the distribution of these particles, enabling unconditional and multi-modal conditional generation.

\subsection{Compositional $k$-Cell Particle Representation}
\label{sec:representation}

A B-Rep model $\mathcal{M}$ is formally defined as a geometric cell complex $\mathcal{C} = \mathcal{V} \cup \mathcal{E} \cup \mathcal{F}$ embedded in $\mathbb{R}^3$. To process this heterogeneous structure uniformly, we abstract $\mathcal{C}$ into its Spatial Hasse Diagram $\mathcal{G} = (\mathcal{N}, \mathcal{I})$, where nodes $\mathcal{N}$ represent the cells (vertices, edges, faces) and directed edges $\mathcal{I}$ represent topological incidence relations (subset inclusion). We define the ``flattened'' representation as the set of nodes from this diagram, resulting in an unordered set of $N$ heterogeneous particles, $\mathcal{P} = \{p_i\}_{i=1}^N$, as illustrated in Figure~\ref{fig:particle_system}. Each particle corresponds to a node $n \in \mathcal{N}$.

\paragraph{Particle Definition.} 
Each particle $p_i = (x_i, c_i, \mathbf{h}_i)$ encapsulates a distinct topological entity:
\begin{itemize}
    \item \textbf{Spatial Anchor $x_i \in \mathbb{R}^3$:} The geometric centroid. For vertices, this acts as the explicit coordinate; for edges and faces, it provides a spatial reference acting as a proxy for the geometric primitive.
    \item \textbf{Cell Type $c_i \in \{0, 1, 2\}$:} The topological dimension (0 for vertex, 1 for edge, 2 for face).
    \item \textbf{Latent Feature $\mathbf{h}_i \in \mathbb{R}^D$:} A learned descriptor encoding the local geometry and, crucially, the topological context derived from the particle's neighbors in the Hasse diagram.
\end{itemize}

Unlike graph-based methods that separate graph structure from feature signal, our particles integrate explicit connectivity into high-dimensional feature correlations. The B-Rep topology is thus represented as a latent graph defined over the particle set, where incidence relationships are explicitly reconstructed during decoding.

\paragraph{Compositional Decoding.}
We employ a compositional decoding scheme aligned with the Hasse diagram structure. The geometry $\Phi_k$ of a $k$-cell ($k>0$) is decoded not solely from its own latent, but as a dependent function of its boundary $(k-1)$-cells $\partial k$:
\begin{equation}
    \Phi_i = \mathcal{D}_{\theta}\big(\mathbf{z}_i, \{\Phi_{j} \mid j \prec_{\mathcal{I}} i\}\big),
\end{equation}
where $j \prec_I i$ denotes that particle $j$ is a boundary of particle $i$ in the recovered incidence graph $\mathcal{I}$. 

\subsection{Compositional Cell VAE (CC-VAE)}
\label{sec:vae}

The CC-VAE, detailed in Figure~\ref{fig:VAE-architecture}, compresses the B-Rep into the latent particle space, preserving the topological hierarchy of the Spatial Hasse diagram within the particle feature space.

\subsubsection{Topology-Aware Encoder}
The encoder (Figure~\ref{fig:VAE-architecture}, Left) aggregates information from dense surface points and topological structures through three distinct levels:

\paragraph{1. Local Geometric Injection.}

Unlike previous primitive-centric encoders, we extract features from the global surface point cloud  $P$ sampled from $\mathcal{M}$. This ensures that the particle aggregates spatially local context beyond its specific boundaries. We employ a cross-attention mechanism where the spatial anchor $x_i$ acts as the query (Q), and the dense cloud $P$ provides keys (K) and values (V). Fourier Positional Encodings (FPE) are applied to retain high-frequency spatial details.

\paragraph{2. Topological Context Aggregation.}
To encode the structural dependencies, we explicitly utilize the input Spatial Hasse Diagram. A 2-layer Graph Convolutional Network (GCN) aggregates features along the incidence links $\mathcal{I}$. The input to the GCN combines the local geometry $f_{\text{local}}^{(i)}$, spatial anchors $x_i$, learned cell type embeddings, Laplacian Positional Encodings (LPE), and the flattened ground truth rotation matrix $\text{flat}(R_i)$ to help pose estimation in decoding stage. The LPEs are crucial for distinguishing spatially adjacent but topologically distinct nodes, explicitly embedding global spectral structural awareness into the latent space.

\paragraph{3. Latent Variational Encoding.}
Finally, the resulting geometry-topology enriched features are processed by a Transformer encoder to capture global dependencies. This maps the set to the variational parameters $(\boldsymbol{\mu}_i, \boldsymbol{\sigma}_i)$ of the posterior distribution $q_\phi(\mathbf{z}_i|\mathcal{M})$, which is regularized via the Kullback-Leibler (KL) divergence against a standard Gaussian prior during training.

\subsubsection{Compositional Geometry Decoder}
Given the latent set $\mathcal{Z}$, the decoder first transforms the particles through self-attention layers into decoded features $\tilde{\mathcal{Z}} = \{\tilde{\mathbf{z}}_i\}_{i=1}^N$. Subsequently, the reconstruction proceeds in two phases: structural recovery and compositional geometric realization, as shown in Figure~\ref{fig:VAE-architecture}.

\paragraph{Phase I: Spatial Hasse Diagram Recovery.}
In this stage, we recover the explicit graph structure. We simultaneously decode two aspects: 
\begin{itemize}[nosep]
    \item \textbf{Entity Attributes.} A projection head predicts the particle type $\hat{c}_i$ and spatial anchor coordinate $\hat{x}_i$, optimized via BCE and L2 losses respectively.
    
    \item \textbf{Incidence Links.} A link prediction head estimates the connectivity probability $A_{ij}$ between any pair via an MLP taking concatenated features $[\tilde{\mathbf{z}}_i, \tilde{\mathbf{z}}_j]$. This effectively reconstructs the edge set $\mathcal{I}$ of the Hasse diagram, trained with a Binary Masked Focal Loss.
    This step establishes a computational graph where every cell explicitly identifies its neighbors for the next phase.
\end{itemize}
    
    \paragraph{Phase II: Compositional Geometric Realization.}
    With the recovered Spatial Hasse Diagram acting as the computational graph, we perform parallel deterministic decoding. Higher-order geometries utilize the reconstructed incidence links to fetch necessary boundary conditions:
    
\begin{itemize}[nosep]
    \item  \textbf{Vertices ($0$-cells).} Defined directly by the predicted anchors $\hat{x}_i$.
    \item  \textbf{Edges ($1$-cells) via Relative Rational Bézier.}
    Edges are modeled as Rational Cubic Bézier curves. To guarantee connectivity, the curve is defined by its two endpoint vertices ($\hat{x}_u, \hat{x}_v$) and two internal control points. An MLP takes the tuple $[\tilde{\mathbf{z}}_u, \tilde{\mathbf{z}}_e, \tilde{\mathbf{z}}_v]$ and predicts the \textit{offsets} and \textit{weights} for the internal control points relative to $\hat{x}_u$ and $\hat{x}_v$. We minimize the $L_1$ loss between sampled points on the predicted curve and the ground truth.
    
    \item \textbf{Faces ($2$-cells) via Canonical Deformation.}
    We generate surface patches by deforming a 2D template within a predicted local canonical frame $T_f = \text{MLP}_{\text{pose}}(\tilde{\mathbf{z}}_f)$. We aggregate the incident particles $\{(x_j, \tilde{\mathbf{z}}_j) \mid j \prec_{\mathcal{I}} f\}$, transform their spatial anchors $x_j$ into the local frame $T_f$, and encode the set using a lightweight PointNet~\cite{PointNet} into a spatial context vector $\mathbf{h}_{\text{spatial}} = \text{PointNet}(\{[T_f(x_j), \tilde{\mathbf{z}}_j] \mid j \prec_{\mathcal{I}} f\})$. The final surface is obtained via a coordinate-based MLP conditioned on both the face latent and the boundary context:
    \begin{equation}
        S(u,v) = T_f \cdot \text{MLP}_{\text{surf}}([u, v, \tilde{\mathbf{z}}_f, \mathbf{h}_{\text{spatial}}]).
    \end{equation}
The module is optimized using a weighted sum of the $L1$ distance for surface samples and a MSE loss for the predicted pose.
\end{itemize}
    
\subsection{Generative Modeling via Rectified Flow}
\label{sec:flow}

With the complex manifold mapped to a concise set $\mathcal{Z}$, we model the generative distribution using Rectified Flow~\cite{RectifiedFlow}. We learn an Ordinary Differential Equation (ODE) that transports a Gaussian prior $\pi_0 = \mathcal{N}(0, I)$ to the data distribution $\pi_1 = p(\mathcal{Z})$ via straight-line paths $z_t = t z_1 + (1-t) z_0$.
 \paragraph{Inference and Assembly.}
We train with a fixed particle budget $N=256$, where input samples with fewer elements are padded via random duplication to reach this capacity. During inference, redundant particles are clustered based on latent space Euclidean distance proximity to recover the precise number of particles.

\begin{figure*}[t]
  \centering
  \includegraphics[width=\linewidth]{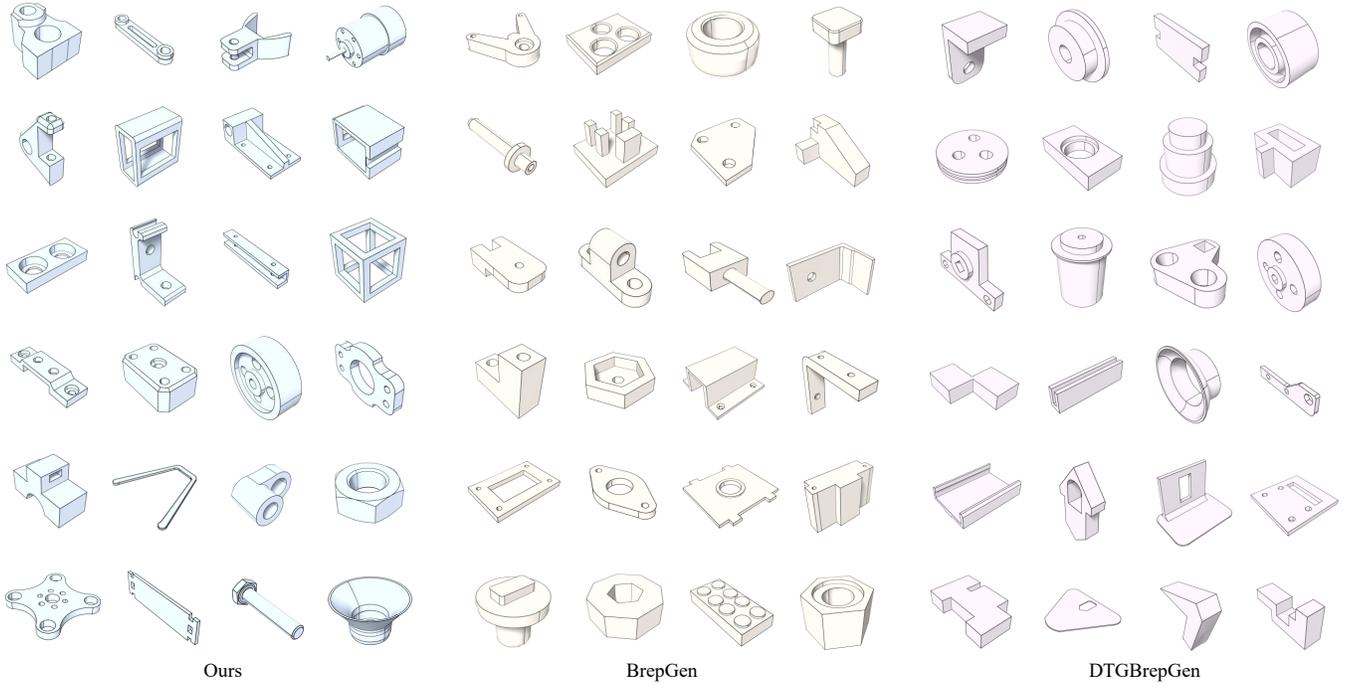}
    \Description{A display of various 3D CAD models generated by the model, showing complex geometries including industrial parts with holes and rounded edges.}
  \caption{Qualitative results of unconditional generation on the ABC dataset after filtering out simple models. The results demonstrate that our method generates complex and realistic CAD models, particularly those featuring intricate details such as chamfers and holes.}
  \vspace{-1em}
  \label{fig:qualitive-results}
\end{figure*}

\begin{table}[h] 
  \centering
  \caption{Quantitative comparison on \textbf{DeepCAD} and \textbf{ABC} datasets. \textbf{Bold} indicates the best result. ($\uparrow$ means higher is better, $\downarrow$ means lower is better).}
  \label{tab:main_results}
  \resizebox{\linewidth}{!}{ 
  \begin{tabular}{l|cccccccc}
    \toprule
    \textbf{Method} & \textbf{1-NNA}$\downarrow$ & \textbf{MMD}$\downarrow$ & \textbf{JSD}$\downarrow$ & \textbf{COV}$\uparrow$ & \textbf{Nov.}$\uparrow$ & \textbf{Uniq.}$\uparrow$ & \textbf{Valid}$\uparrow$ & \textbf{CC}$\uparrow$ \\
    \midrule
    \multicolumn{8}{c}{\textbf{DeepCAD Dataset}} \\
    \midrule
    
    BRepGen \cite{BrepGen} 
    & 64.37 & 1.40 & 1.47 & 68.94 & \textbf{99.03} & 99.93 & 59.84 & 8.90 \\

    DTG-BRepGen \cite{DTGBrepGen} 
    & \textbf{57.46} & \textbf{1.38} & \textbf{0.76} & \textbf{70.41} & 85.82 & 98.86 & \textbf{87.23} & 9.04\\
    
    \rowcolor{gray!10} 
    \textbf{Ours} 
    & 60.58 & 1.38 & 1.49  & 68.41 & 94.71 & \textbf{99.64} & 86.68 & \textbf{11.21}\\

    \midrule
    \multicolumn{8}{c}{\textbf{ABC Dataset}} \\
    \midrule

    BRepGen \cite{BrepGen} 
    & 67.61 & 1.84 & 2.44 & 63.02 & 98.75 & 99.68 & 40.32 & 10.51 \\

    DTG-BRepGen \cite{DTGBrepGen} 
    & 63.40 & 1.81 & 1.13 & 63.98 & 89.86 & 98.00 & 62.14 & 9.78 \\
    
    \rowcolor{gray!10} 
    \textbf{Ours} 
    & \textbf{63.02} & \textbf{1.74}& \textbf{0.66} & \textbf{64.32} & \textbf{97.60} & \textbf{99.93} & \textbf{66.50} & \textbf{12.92} \\
    
    \bottomrule
  \end{tabular}
  }
\end{table}

  \section{Implementation Details}

	\subsection{Architecture and Training}

  \paragraph{CC-VAE} 
  The encoder and decoder comprise 4 and 8 self-attention layers, respectively, with a model dimension of 512. The latent embedding is compressed to a dimension of 16 to facilitate efficient diffusion modeling. The KL divergence loss is set to $2\times 10^{-6}$. The VAE is trained with a learning rate of $1\times 10^{-4}$ until the reconstruction loss converges. 

  \paragraph{Flow Matching Backbone} 
  We adopt a 12-layer Transformer backbone with a hidden dimension of 768. We incorporate RMSNorm and SwiGLU activation functions to stabilize training~\cite{SD3, SwiGLU}. We optimize the Flow Matching objective using AdamW ($\beta_1=0.9, \beta_2=0.95$)~\cite{BETA95} with a batch size of 1,080 distributed across 6 NVIDIA RTX 4090 GPUs. Key training strategies include: (i) an Exponential Moving Average (EMA) decay of 0.9999; (ii) logit-normal $t$-sampling ($m=0, s=1$)\cite{SD3}; and (iii) a velocity direction loss~\cite{FasterDiT}. The unconditional model is trained for 300k iterations. 

  For conditional generation tasks, we adopt a Dual-Stream architecture inspired by MM-DiT~\cite{SD3}. Conditional tokens and noisy latent tokens are processed in parallel streams. A joint attention mechanism facilitates bidirectional information flow between streams without collapsing their feature spaces into a single sequence, ensuring effective multi-modal alignment. We initialize the backbone with weights pre-trained on the ABC dataset and fine-tune the model on DeepCAD and Roof dataset for additional 100k iterations. 
  
  For the in-painting task, we simulate partial constraints by randomly selecting 20\

  \subsection{Post-Processing}Geometric primitives are fitted to the decoded geometry with a preference for analytical shapes over freeform surfaces, and assembled according to the decoded topology. After resolving loop orientation via a largest-area heuristic, the final model is reconstructed as a B-Rep using OCCT~\cite{OCCT}.

  \section{Experimental Results}
  
  We validate our method on three key dimensions: generation quality on standard benchmarks, topological robustness on complex geometries, and versatility in downstream editing tasks.
  
  \subsection{Experimental Setup} 
  \paragraph{Datasets.}
  We evaluate our method using the ABC~\cite{ABC}, DeepCAD~\cite{DeepCAD}, and Furniture~\cite{BrepGen} datasets. Furthermore, to validate generalization in real-world scenarios, we utilize the Roof dataset~\cite{ROOF}, comprising planar 3D polygonal roof meshes paired with aerial imagery.
  
  Consistent with previous protocols~\cite{DTGBrepGen, SolidGen}, we apply data filtering to remove overly complex or duplicated samples. Specifically, we exclude models with more than 50 faces, more than 30 edges per face, or a total cell count ($V+E+F$) exceeding 192. This filtering process results in 188,747 samples for ABC, 92,013 for DeepCAD, 1,624 for Furniture, and 3,584 for the Roof dataset. All datasets are randomly split into training, testing, and validation sets with a ratio of 90\
  
  \paragraph{Evaluation Protocol and Metrics.}
  Following previous work~\cite{BrepGen, SolidGen}, we generate 3,000 samples from Gaussian noise and randomly select 1,000 reference samples from the test set for each metric run, and average metrics over 10 independent runs.
  
  We adopt a comprehensive set of standard metrics~\cite{BrepGen, SolidGen} to assess the performance of our model from three perspectives: (1) \textit{Distribution Quality} is evaluated using \textbf{Minimum Matching Distance (MMD)} for geometric fidelity, \textbf{Jensen-Shannon Divergence (JSD)} for point distribution discrepancy, and \textbf{1-NNA} for indistinguishability. (2) \textit{Diversity} is measured via \textbf{Coverage (COV)}, \textbf{Novelty}, and \textbf{Uniqueness} to ensure the generation of varied samples. (3)  \textit{CAD Quality} focuses on structural validity and complexity, reporting the standard \textbf{Valid Rate} for watertightness, \textbf{Min-$k$ Validity} to assess robustness across varying geometric densities, and \textbf{Cyclomatic Complexity (CC)}~\cite{CC} for topological intricacy.
  
  \subsection{Unconditional Generation on Solids}

  We first evaluate the capability of our model to synthesize high-quality B-Reps from Gaussian noise, benchmarking against state-of-the-art direct B-Rep generation methods: BRepGen~\cite{BrepGen}, and DTG-BRepGen~\cite{DTGBrepGen}. As shown in Figure~\ref{fig:qualitive-results}, our method produces visually plausible samples that exhibit high geometric fidelity and structural coherence.
  
  \paragraph{Quantitative Results on Standard Benchmarks.} On the ABC dataset, our method demonstrates a decisive advantage by achieving a remarkable balance between quality and complexity. Specifically, we attain a Validity score of 66.50\

\begin{figure}[t]
    \centering
    \includegraphics[width=\linewidth]{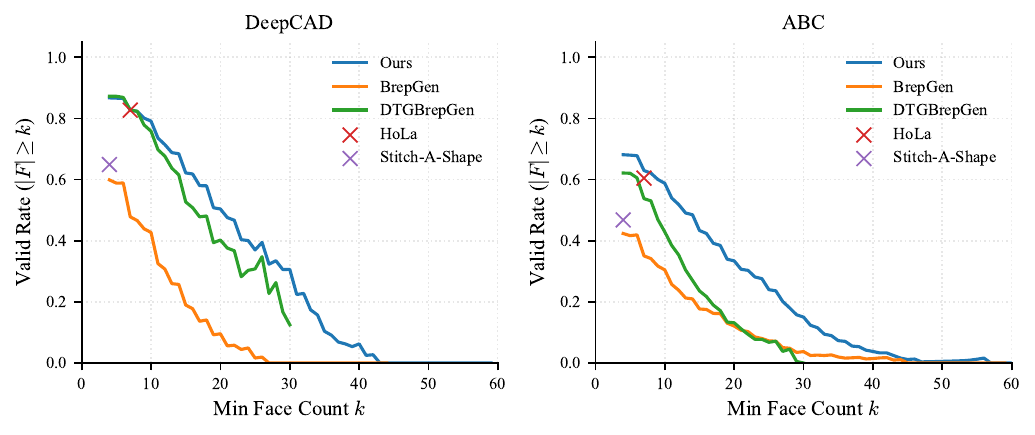}
    \Description{Line graph comparing the validity rate of different methods as the minimum number of faces increases.}
    \caption{We plot the validity rate of generated models as a function of the minimum number of faces ($\min\text{-}k$).  Markers 'x' indicate performance points reported in HoLa~\cite{HoLa} and Stitch-A-Shape~\cite{StitchAShape}.}
    \vspace{-1em}
    \label{fig:complexity_curve}
    \end{figure}

  \paragraph{Robustness against Geometric Complexity.}
  Global aggregate metrics often mask performance nuances, particularly regarding the ability to generate non-trivial shapes. To rigorously investigate capability on complex designs, we evaluate the \textit{min-$k$ Valid Rate}, which measures the validity of generated samples containing at least $k$ faces.
  As visualized in Figure~\ref{fig:complexity_curve}, our method demonstrates superior robustness. Most baseline methods exhibit a decay as $k$ increases, and a precipitous drop in validity at $k=7$. Since a standard cuboid consists of 6 faces ($k=6$), this sharp decay suggests that the average valid rates of prior works are partially inflated by trivial box-like shapes.
  Our method, however,  consistently outperforms all baselines as  $k$ increases. This confirms that our unified generation paradigm effectively disentangles topological dependencies, enabling the successful synthesis of complex engineering parts that typically baffle cascaded models.

    \begin{figure}[t]
      \centering
      \includegraphics[width=\linewidth]{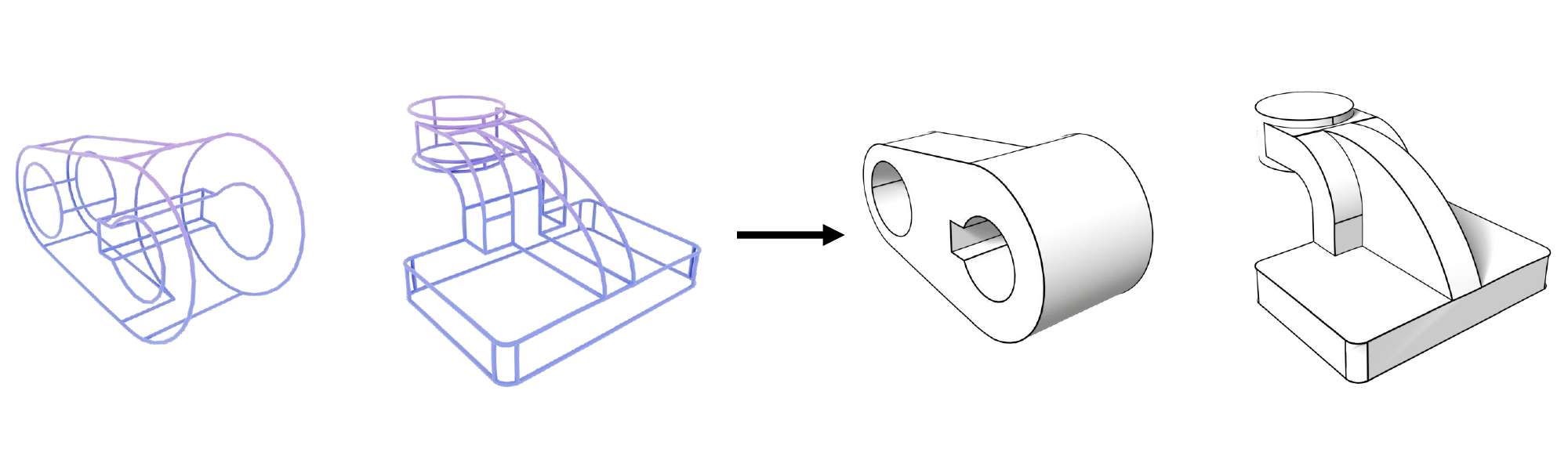}
      \Description{Inpainting  results.}
      \caption{Given only the edge and vertex constraints,the model successfully predicts the missing faces to produce the complete solid B-Reps.}
      \vspace{-1em}
      \label{fig:inpainting}
    \end{figure}
  
  \subsection{Conditional Generation and Editing}
  
  Our unified particle representation naturally supports multi-modal conditioning and partial editing tasks without architectural changes.
  
  \paragraph{Single-view and Point Cloud Reconstruction.}
  We evaluate conditional synthesis on the DeepCAD dataset~\cite{DeepCAD}.
  
  (1) \textbf{Image-Conditioned:} Utilizing pre-trained DINOv2~\cite{DINOv2} embeddings, we reconstruct 3D B-Reps directly from single-view renderings or aerial images.
  
  (2) \textbf{Point Cloud-Conditioned:} Using pre-trained Sonata~\cite{Sonata} to encode sparse 3D data, we generate complete B-Reps from sampled 2048-point clouds.

  Qualitative results in Figure \ref{fig:conditioned-cad} demonstrate our generated models reflect the input geometry with high fidelity. When generating multiple samples from a single input, we observe minimal variance among results; stochastic variations are strictly confined to inherently ambiguous regions such as occluded surfaces in single-view images or fine-grained details. This stability indicates that our model captures the precise, explicit topology of the input,  rather than merely reconstructing a coarse global approximation.

  \paragraph{Local In-painting}
  The explicit, localized nature of our particle set allows for masked generation. We demonstrate this via a Wireframe-to-Shape task. By fixing the tokens corresponding to edges/vertices and masking face tokens, we essentially in-paint the surfaces. See \ref{fig:inpainting}, Our model successfully recovers watertight surfaces with correct shapes given only the skeletal wireframe, proving its capacity for local context completion and interactive CAD editing. 

  \begin{figure}[t]
    \centering
      \includegraphics[width=\linewidth]{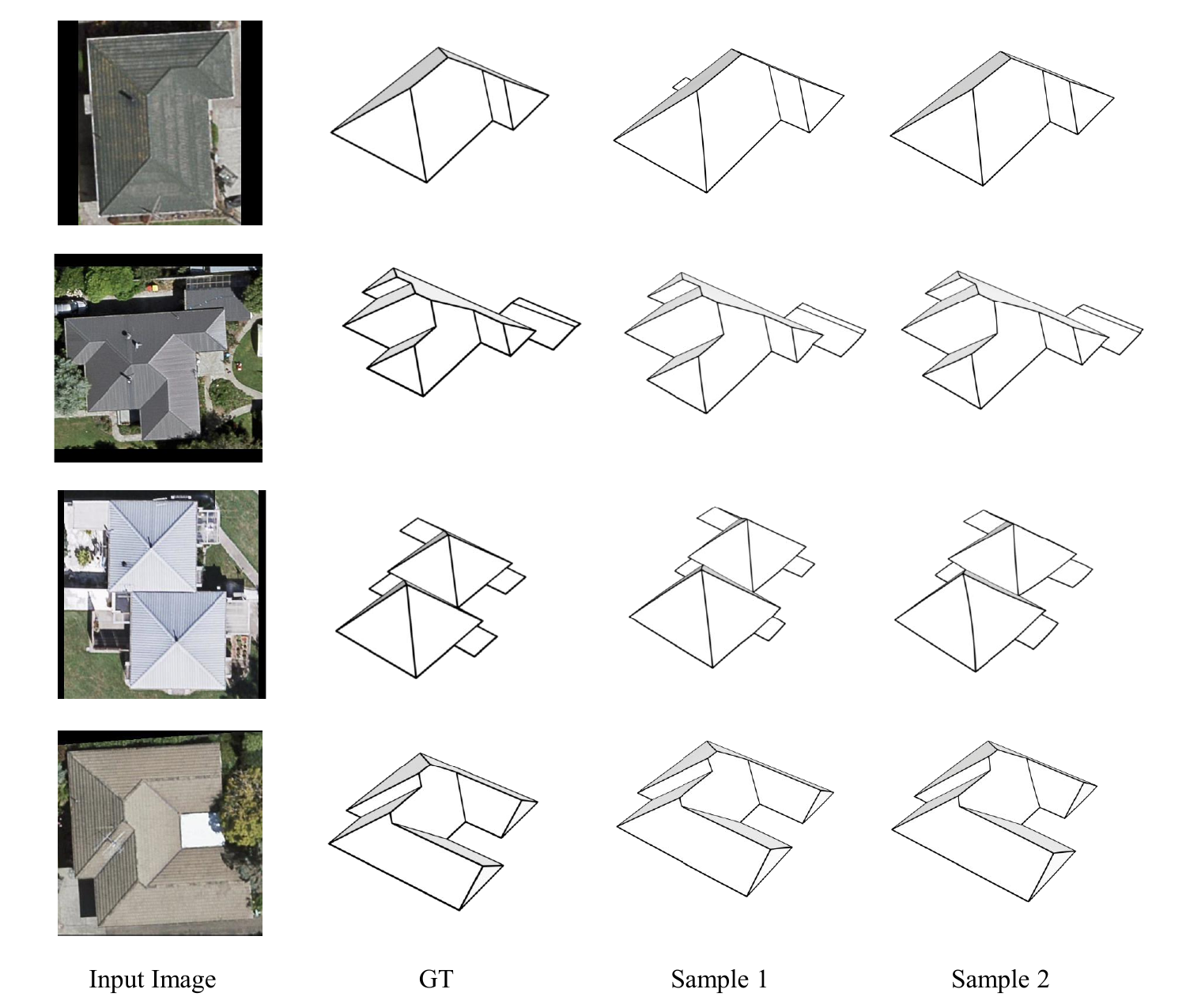}
    \Description{Given aerial photographs as input, we generate polygonal roof meshes.}
    \caption{\textbf{Reconstruction from Real-world Satellite Imagery.} 
  Given aerial photographs as input, our model generates clean, polygonal roof meshes. }
      \vspace{-1em}
    \label{fig:roof-results}
  \end{figure}
  
  \subsection{Generalization to Non-Solid Topologies}

  \paragraph{Satellite Roof Reconstruction.}
  We assess robustness on open surfaces using the Roof dataset \cite{ROOF}, which consists of planar meshes derived from satellite imagery. 
  As shown in Figure \ref{fig:roof-results}, our method generalizes effortlessly to this domain, accurately reconstructing ridge lines and planes. This confirms that our generative prior is not limited to closed solids but extends to generic $k$-cell complexes suitable for GIS applications.
  
\begin{figure*}[p] 
    \centering
    
    \includegraphics[width=\linewidth]{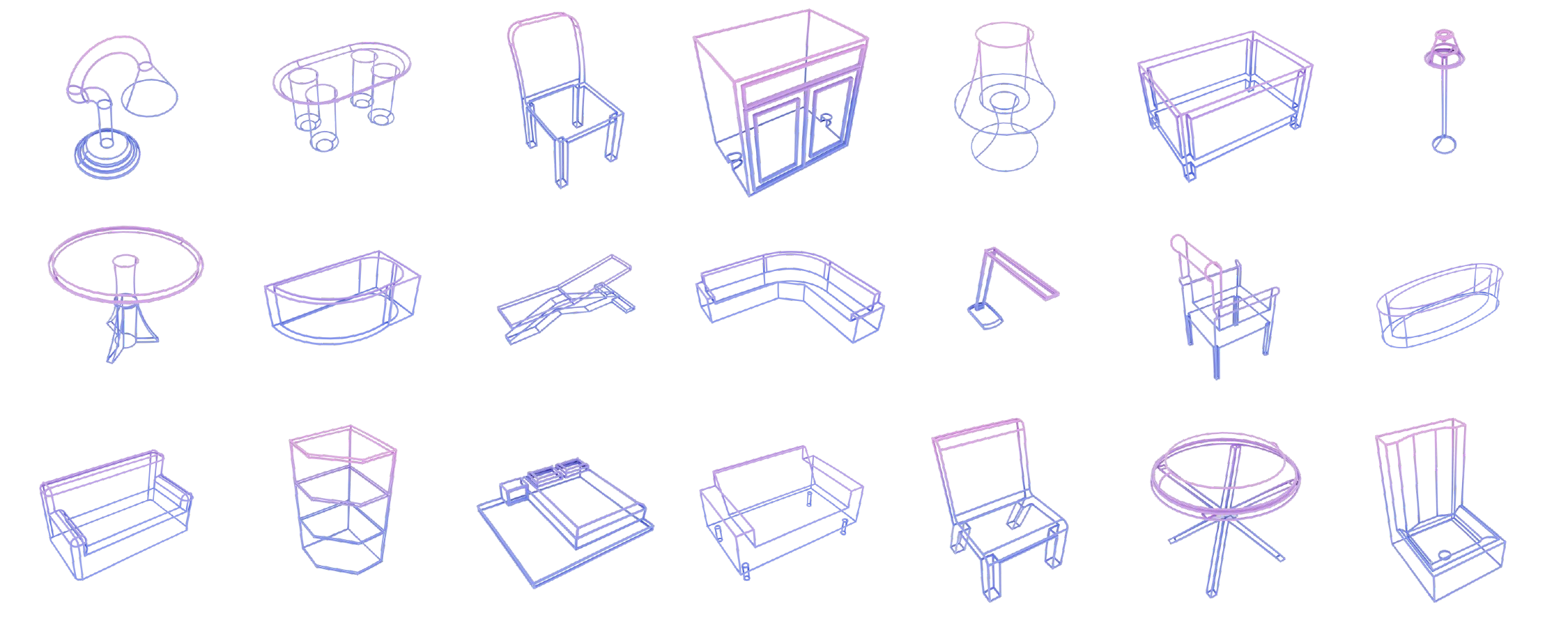}
    \Description{Generated wireframe models exhibiting complex topologies and curves.}
      \caption{\textbf{Unconditional Generation on Furniture Wireframes.} Our model is capable of synthesizing the 1-skeleton of shapes, handling both straight structural lines and complex organic curves (e.g., chair backs and legs) with high fidelity, demonstrating the flexibility of our particle representation beyond watertight solids.}
    \label{fig:wireframe-generation}
    
    \vspace{1cm}

    \includegraphics[width=\linewidth]{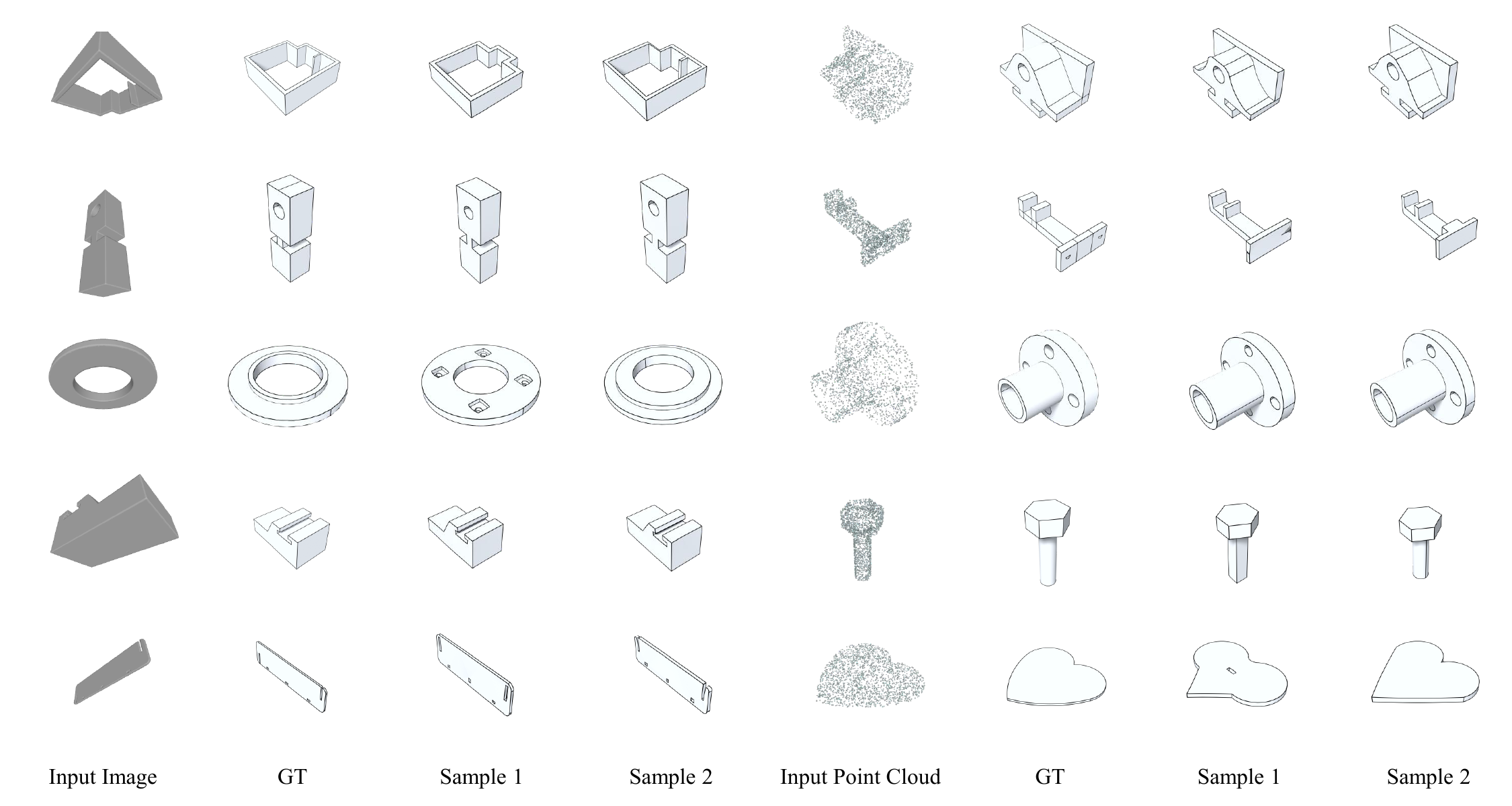}
    \Description{[TODO] A detailed description of the image content for accessibility.}
    \caption{\textbf{Multi-modal Conditional Generation.} 
    We demonstrate shape reconstruction performance across different input modalities. 
    \textbf{Left:} Reconstruction from single-view images. 
    \textbf{Right:} Reconstruction from sparse point clouds. 
    These results confirm that our unified particle formulation naturally facilitates cross-modal alignment, producing high-fidelity B-Reps that accurately adhere to diverse input signals.}
    \label{fig:conditioned-cad}
\end{figure*}

  \paragraph{Unconditional Wireframe Synthesis.}
  To further evaluate topological flexibility, we train exclusively on wireframes of the Furniture dataset \cite{BrepGen}. Figure \ref{fig:wireframe-generation} shows that our model synthesizes complex furniture structures without architectural changes. It effectively captures both structural straight lines and the organic, free-form splines typical of furniture design. This implies that our particle-based formulation is a general-purpose geometric framework, capable of modeling arbitrary connectivity beyond the algebraic constraints of mechanical parts.
  
  \begin{table}[h]
	  \centering
	  \caption{Increasing the number of initial tokens at test time directly improves topological validity and model complexity coverage, demonstrating strong extrapolation capabilities.}
	  \label{tab:inference_scaling}
	  \resizebox{0.8\linewidth}{!}{
	  \begin{tabular}{l|ccc}
		  \toprule
		  \textbf{Particle Count} & \textbf{DeepCAD Validity} $\uparrow$ & \textbf{ABC Validity} $\uparrow$ \\
		  \midrule
		  256           & 68.64 & 51.32 \\
		  512		    & 78.65 & 62.98 \\
		  1024          & \textbf{86.68} & \textbf{66.50} \\
		  \bottomrule
	  \end{tabular}
	  }
   \vspace{-1em}
  \end{table}

  \subsection{Inference-Time Scaling.}
	\label{sec:ablation}

  A unique property of our unordered, particle-based formulation is its ability to extrapolate. Although trained with a fixed upper bound of nodes of 256, our architecture as a set transformer can accept varying numbers of latent particles at inference.
  Table~\ref{tab:inference_scaling} reveals a compelling trend: simply increasing the number of noise particles at $t=1$ significantly boosts the validity and fidelity of the generated B-Reps. This "scaling law" suggests our model learns a continuous geometric field rather than a fixed discrete template, allowing users to trade off compute for higher fidelity without any retraining which is a capability largely absent in fixed-sequence autoregressive baselines.

  \begin{figure}[t]
    \centering
    \includegraphics[width=0.9\linewidth]{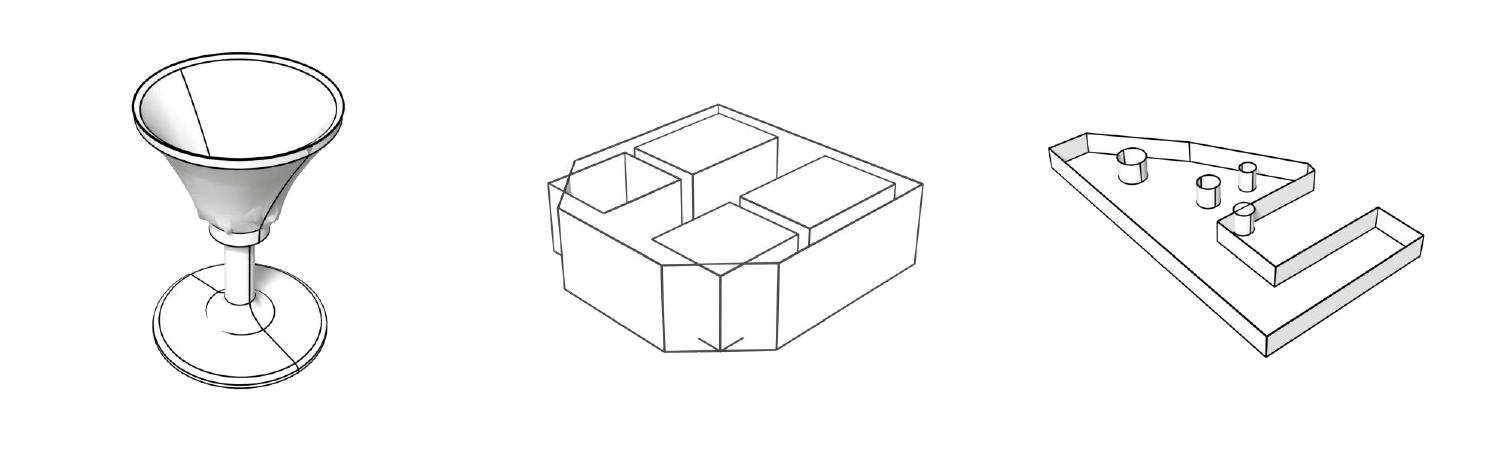}
    \Description{Three 3D models showing failure cases. Left: a chalice with a wrinkled surface. Middle: a boxy structure with intersecting walls. Right: a mechanical part with overlapping features.}
    \caption{\textbf{Failure Cases} Left: Geometric artifacts on a watertight surface
    Middle \& Right: Invalid geometry due to geometric interpenetration of topologically disjoint parts.}
    \vspace{-1em}
    \label{fig:failure}
  \end{figure}
  
  \section{Limitation and Conclusion}While our method demonstrates strong performance, we acknowledge three limitations. First, geometric anomalies such as wrinkled faces or boundary violations (Figure~\ref{fig:failure}) can still occur even in topologically valid samples. Second, explicitly encoding all $k$-cells yields a larger latent space than primitive-based methods, increasing computational overhead and potentially constraining scalability for dense shapes. Third, while strict geometric constraints are theoretically compatible with our framework, our primary focus in this work is validating the topological expressiveness and generation capability of the Compositional k-Cell Particles representation. The integration of differentiable analytic solvers introduces distinct optimization challenges that we consider a separate avenue of research.
  
  In conclusion, we introduced a novel framework that reformulates B-Reps as sets of compositional $k$-cell particles. By unifying geometry and topology via flow matching, our method achieves state-of-the-art validity and complexity coverage. Extensive experiments confirm its versatility across multi-modal conditioning, wireframe synthesis, and open-surface reconstruction, paving the way for more flexible and robust generative CAD systems.

% \input{sections/conclusion}

%% Bibliography
\bibliographystyle{ACM-Reference-Format}
\bibliography{main}

\newpage
\clearpage

\newpage
\begin{figure*}[p]
    \centering
    \includegraphics[width=\linewidth]{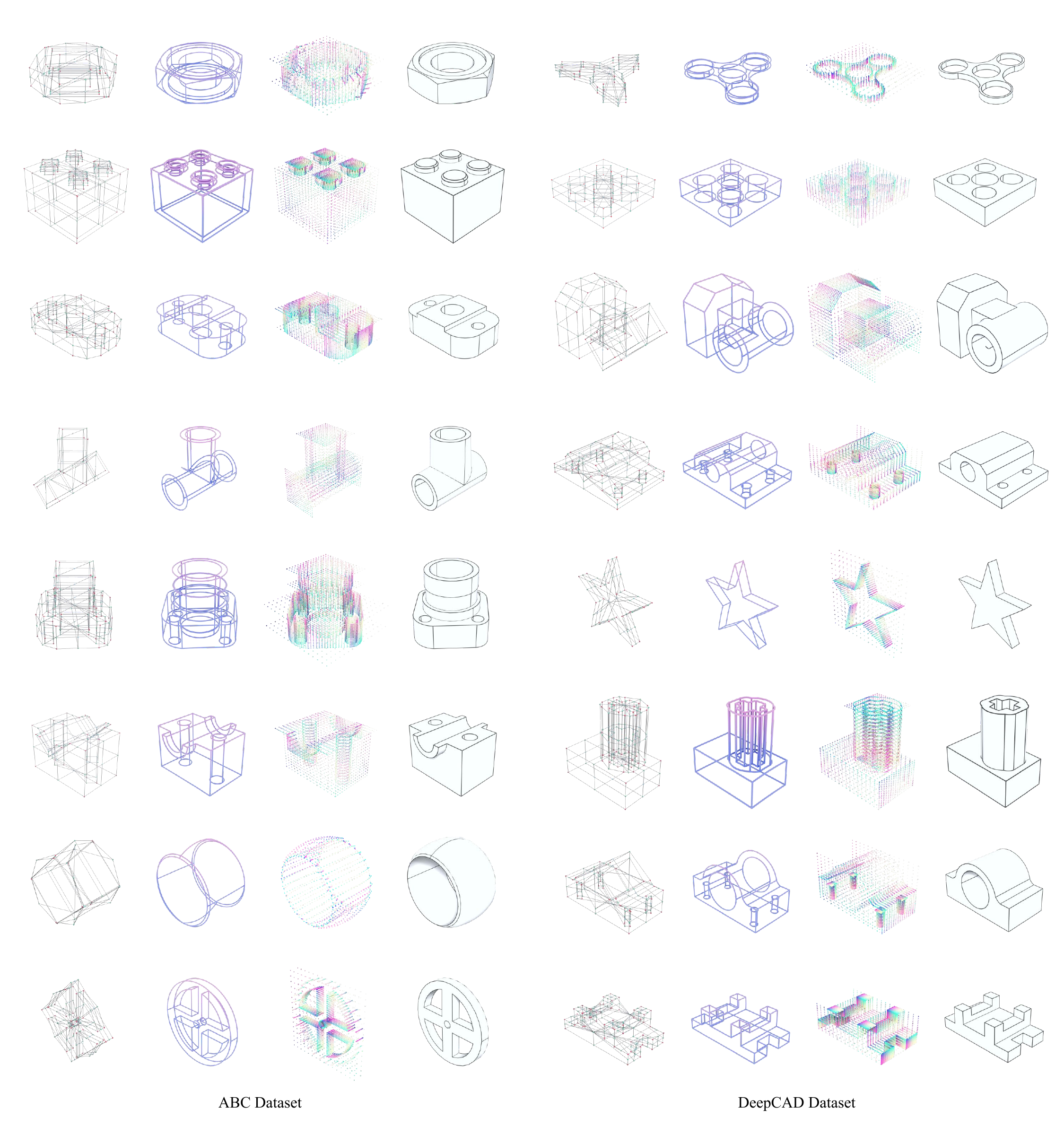}
    \Description{[todo]}
    \caption{\textbf{Extended Generation Gallery.} More unconditional generation results on ABC Dataset(left) and DeepCAD Dataset (right) datasets. Each entry displays (from left to right): the Spatial Hasse Diagram, edge geometry, face geometry, and the final B-Rep model. }
    \label{fig:more-results}
\end{figure*}

\end{document}